\documentclass[letterpaper, 10 pt, conference]{ieeeconf}  % Comment this line out if you need a4paper

\IEEEoverridecommandlockouts                              % This command is only needed if 
\usepackage{graphics} % for pdf, bitmapped graphics files
\usepackage{epsfig} % for postscript graphics files
\usepackage{mathptmx} % assumes new font selection scheme installed
\usepackage{times} % assumes new font selection scheme installed
\usepackage{amsmath} % assumes amsmath package installed
\usepackage{amssymb}  % assumes amsmath package installed
\usepackage{xcolor}

\title{\LARGE \bf
Memory-based Semantic Segmentation for Off-road Unstructured Natural Environments
}

\author{Youngsaeng Jin$^{1}$, David Han$^{2}$ and Hanseok Ko$^{1}$% <-this % stops a space
\thanks{*This work was supported by Air Force Office of Scientific Research under award number FA2386-19-1-4001.}% <-this % stops a space
\thanks{$^{1}$Youngsaeng Jin and Hanseok Ko are with the School of Electrical Engineering, Korea University, Seoul 136-713, South Korea
        {\tt\small youngsjin@korea.ac.kr; hsko@korea.ac.kr}}%
\thanks{$^{2}$David K. Han is with Electrical and Computer Engineering, Drexel University, Philadelphia, PA 19104 {\tt\small dkh42@drexel.edu}}%
}

\begin{document}

\maketitle
\thispagestyle{empty}
\pagestyle{empty}

%%%%%%%%%%%%%%%%%%%%%%%%%%%%%%%%%%%%%%%%%%%%%%%%%%%%%%%%%%%%%%%%%%%%%%%%%%%%%%%%
\begin{abstract}

With the availability of many datasets tailored for autonomous driving in real-world urban scenes, semantic segmentation for urban driving scenes achieves significant progress. However, semantic segmentation for off-road, unstructured environments is not widely studied. Directly applying existing segmentation networks often results in performance degradation as they cannot overcome intrinsic problems in such environments, such as illumination changes. In this paper, a built-in memory module for semantic segmentation is proposed to overcome these problems. The memory module stores significant representations of training images as memory items. In addition to the encoder embedding like items together, the proposed memory module is specifically designed to cluster together instances of the same class even when there are significant variances in embedded features. Therefore, it makes segmentation networks better deal with unexpected illumination changes. A triplet loss is used in training to minimize redundancy in storing discriminative representations of the memory module. The proposed memory module is general so that it can be adopted in a variety of networks. We conduct experiments on the Robot Unstructured Ground Driving (RUGD) dataset and RELLIS dataset, which are collected from off-road, unstructured natural environments. Experimental results show that the proposed memory module improves the performance of existing segmentation networks and contributes to capturing unclear objects over various off-road, unstructured natural scenes with equivalent computational cost and network parameters. As the proposed method can be integrated into compact networks, it presents a viable approach for resource-limited small autonomous platforms.

\end{abstract}

%%%%%%%%%%%%%%%%%%%%%%%%%%%%%%%%%%%%%%%%%%%%%%%%%%%%%%%%%%%%%%%%%%%%%%%%%%%%%%%%
\section{INTRODUCTION}

Semantic segmentation is a fundamental but significant task for scene understanding. It aims to assign semantic labels for each pixel in the image. As semantic segmentation provides diverse information including categories, locations, and shapes of objects, it is critical to a variety of real-world applications, such as robot vision~\cite{asadi2019real}, autonomous driving~\cite{siam2018comparative}, medical diagnosis~\cite{zhao2013overview}, etc. However, this task is challenging to achieve high accuracy due to label and shape variety.

With the developments of deep neural networks, Shelhamer \textit{et al.}~\cite{fcn} proposed the Fully Convolutional Network (FCN), which attained an impressive improvement on semantic segmentation accuracy. Due to the effectiveness of the FCN, it is employed as a core framework in the state-of-art semantic segmentation methods. Some efforts~\cite{pspnet,deeplab,deeplabv3} aggregate multi-scale contextual information to capture multi-scale objects. Other efforts~\cite{zhao2018psanet,danet,ccnet} use an attention mechanism~\cite{bahdanau2014neural,xu2015show} to capture richer global contextual information.

Another factor making progress in the development of semantic segmentation is the availability of semantic segmentation datasets~\cite{pascalcontext,ade20k}. These datasets were collected from various real-world environments and have been provided for advancing the technique in real-world applications. In particular, several datasets tailored for autonomous driving in real-world urban scenes~\cite{kitti,camvid,cityscapes,huang2019apolloscape} have been provided and exploited as a primary source of data for navigating autonomous vehicles. As the urban scenes are structured environments with low variations of scenes and illumination, they are relatively easier to be segmented precisely. Thus, a significant development for autonomous driving in urban environments has been achieved.

\begin{figure}[t]
\centering
\includegraphics[width=0.45\textwidth]{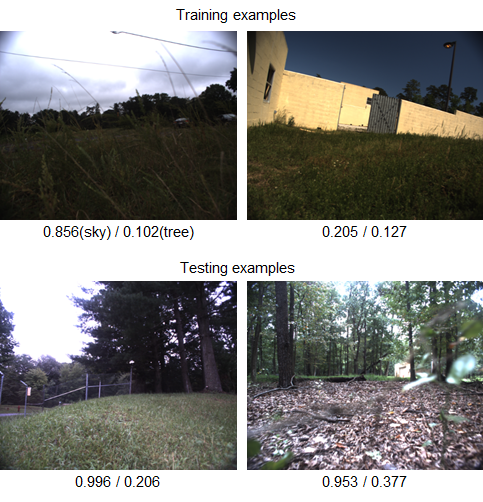}
\caption{Image samples from RUGD dataset~\cite{RUGD2019IROS}, collected from off-road, unstructured environments. The numbers below each image are the average pixel intensity of categories `sky' and `tree'. Images in this dataset cover a wide range of scenes and their illumination is inconsistent.}
\label{fig:exampleImg}
\end{figure}

\begin{figure}[t]
\centering
\includegraphics[width=0.46\textwidth]{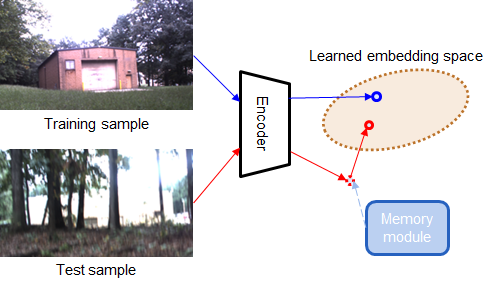}
\caption{\textcolor{black}{The memory module shifts representations of samples with significant variances into the learned embedding space so that the overall segmentation networks better deal with the unexpected illumination changes.}}
\label{fig:ideaMem}
\end{figure}

However, when navigating in off-road, unstructured natural environments, an autonomous platform encounters formidable challenges to recognize its surroundings and objects therein. Such scenarios face not only a wide range of scenes but also significant illumination changes as some examples shown in Fig.~\ref{fig:exampleImg}. Unfortunately, as there are many factors, such as camera sensitivity or lighting conditions, causing the illumination changes, these changes are unavoidable during navigating. At worst, an object becomes as brighter (or darker) as surrounding regions and it is hard to distinguish from them as the building in the `Test sample' of Fig.~\ref{fig:ideaMem}. Therefore, as illumination plays a critical role in capturing the appearance, inconsistent illumination results in performance degradation.
% \textcolor{black}{In addition, as collecting a large number of image-label pairs for semantic segmentation is difficult and time-consuming, it is often incapable of generalizing to cover a wide range of scenes using limited quantity of samples.}

Motivated by the above issues, in this paper, we propose a built-in memory module for semantic segmentation to improve semantic segmentation accuracy for off-road, unstructured natural environments. The memory module stores the significant representations of training images as memory items. Then, the memory items are recalled to cluster the instances of the same class closer within the embedding space learned by training images. Therefore, the memory module mitigates significant variances in embedding features and segmentation networks with the memory module better deal with the unexpected illumination changes as illustrated in Fig.~\ref{fig:ideaMem}. In our experimental configuration, the proposed memory module follows after an encoder to refine the global contextual features (encoder output feature maps) using memory items. Then, the decoder takes the refined features and produces the segmentation masks. In order not to affect the computational cost, the memory module contains a few items. In addition, the triplet loss~\cite{tripletloss2} is used to make the items far apart and minimize the redundancy of the items. Our proposed memory module is very general thus can be adopted on a wide range of networks.

Experiments are conducted on the Robot Unstructured Ground Driving (RUGD) dataset~\cite{RUGD2019IROS} and RELLIS~\cite{jiang2020rellis3d} dataset, which are collected by an unmanned ground robot from off-road, unstructured natural environments. The quantitative results show that the proposed memory module improves the performance of existing segmentation networks with equivalent computational cost and network parameters regardless of compact or complex networks. The qualitative results demonstrate the effectiveness in capturing unclear objects over a variety of off-road, unstructured scenes. Our memory module applied to compact segmentation networks delivers improved performance on outdoor scene segmentation in real-time operation, thus it allows better autonomous navigation for resource-limited small autonomous platforms.

% The rest of the paper is structured as follows. Section II introduces the related works. Section III presents the proposed method. Section IV gives the experimental results with analysis, and Section V concludes our method.

\section{Related Work}

\subsection{Semantic Segmentation}

Driven by the development of Convolutional Neural Networks (CNNs), current semantic segmentation methods always employ deep CNNs (e.g., ResNet~\cite{resnet}, ResNext~\cite{resnext}, etc.) as an encoder to extract feature representations. To improve the performance, various decoder modules are proposed to produce precise segmentation masks. 

A variety of early methods~\cite{fcn,vemulapalli2016gaussian,deeplab,deconvnet,unet} have been proposed, but these approaches often resulted in poor performance due to their model simplicity.
To boost performance, more advanced methods have been developed. PSPNet~\cite{pspnet} and Deeplab~\cite{deeplabv3,deeplabv3p} frameworks incorporated multi-scale contextual information using spatial pyramid module while some methods~\cite{bilinski2018dense,denseaspp, gff} used dense layers in the decoder end to infuse multi-level features for advantage of feature reuse.
Some efforts~\cite{danet,yuan2018ocnet,zhang2018context,zhao2018psanet,ccnet,ANNN,jin2021trseg} exploited attention mechanisms to capture long-range dependency for richer global contextual information.
In addition, a gate mechanism~\cite{gru} is employed to selectively fuse multi-level features or multi-modal features to further improve the performance~\cite{depthseg,GMA,gscnn,gff,jin2021sketch}.
Effectiveness of these methods has been verified on datasets collected from structured environments (e.g., Cityscapes~\cite{cityscapes}, ADE20K~\cite{ade20k}, etc.), but their effectiveness on an off-road, unstructured environment has not yet been verified.

\subsection{Memory Networks}
Graves \textit{et al.}~\cite{graves2014neural} introduced a concept called "Neural Turing Machine", which combines neural networks with an external memory bank to extend neural networks capability. The external memory is jointly trained with the main branch. The combined architecture uses an attention process to selectively read from and write to the memory. Due to its flexibility, it is adapted to a variety of tasks, such as few-shot learning~\cite{santoro2016meta,kaiser2017learning}, video summarization~\cite{lee2018memory}, image captioning~\cite{chunseong2017attend}, anomaly detection~\cite{mem_abnomaly}, etc.
In our method, a memory module is exploited for semantic segmentation to deal with unexpected illumination changes in an off-road, unstructured environment by storing the significant representations of training images and recalling the representations to correct significant variances in embedded features.

% The significant representations within a training set are recorded as memory items, and the items are recalled during the testing phase to embed exceptional regions of testing images into the embedding space of training images to overcome the.

\section{Method}

In this section, we first provide an overview of the semantic segmentation framework. Then we present the proposed memory module and the loss function.

\begin{figure}[t]
\centering
\includegraphics[width=0.48\textwidth]{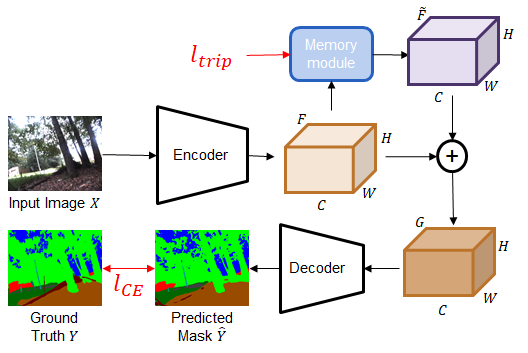}
\caption{Illustration of the overall architecture. \(l_{CE}\) and \(l_{trip}\) denote the cross entropy loss and the triplet loss, respectively.}
\label{fig:overall_archi}
\end{figure}

\subsection{Overview of Semantic Segmentation Framework}

As deep convolutional neural networks are capable of extracting salient features from input images, they are employed as an encoder. To alleviate the loss of the details, they produce high-resolution feature maps to preserve information. It is implemented by replacing some of the last stride convolutions~\cite{kong2017take,drn} with dilated convolutions~\cite{yu2015multi} correspondingly. The ratio of input image spatial resolution to the encoder output resolution, denoting as output stride (OS), is often 8 or 16\footnote{Lower OS improves the segmentation accuracy but requires more computational cost.}.
% As such, Dilated Residual Networks (DRN)~\cite{drn} are often employed as the encoder. 
Starting with the input image \(X\in R^{3\times H_0\times W_0}\) whose height and width are \(H_0\) and \(W_0\), an encoder produces the global contextual feature maps \(F\in R^{C\times H\times W}\), where \(C, H, W\) are the number of channels, height and width respectively, at the final layer.
Then a decoder takes the feature maps as input to produce a segmentation mask. It is often a sub-network, such as Atrous Spatial Pyramid Pooling (ASPP) module in Deeplabv3~\cite{deeplabv3}, to refine the feature maps for the performance improvement and produce the prediction map. At last, the prediction map is bilinearly upsampled to the resolution of the input image for the final segmentation result \(\hat{Y}\in R^{N_{cls}\times H_0\times W_0}\), where \(N_{cls}\) is the number of the categories. Although the decoder often improves the performance, they are ineffective if the encoder output feature maps do not properly represent the input image. To mitigate this issue, we propose a memory module to refine the feature maps using the memory items before feeding them into the decoder as depicted in Fig.~\ref{fig:overall_archi}. The details of the memory module are presented in the following subsection.

\subsection{Memory Module}

\begin{figure}[t]
\centerline{\includegraphics[width=0.48\textwidth]{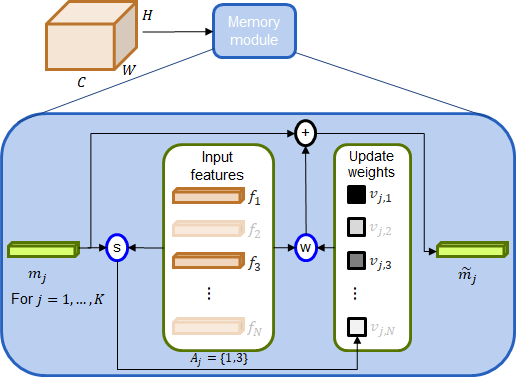}}
\caption{Illustration of the process of memory write. Given the \(j^{th}\) memory item \(m_j\), we look for a set \(A_j\) containing indexes of individual features which have highest addressing weight on \(m_j\) as in (5). Then we compute the update weights in (6)$\sim$(8) and the memory item is updated by a weighted sum of the individual features in \(A_j\) with the corresponding update weights as shown applied in (9). \textcircled{\raisebox{-.9pt} {$\textup{+}$}} is the element-wise summation. \(A_j=\{1,3\}\) in this example.}
\label{fig:mem_update}
\end{figure}

The memory module performs read and write operations. The read operation refines encoder output feature maps using stored memory items while the write operation updates the memory items according to the encoder output feature maps. The write operation is only conducted during the training phase. The read operation (Fig.~\ref{fig:mem_read}) is presented first and then the write operation is followed (Fig.~\ref{fig:mem_update}).

\subsubsection{Read} 

Given encoder output feature maps \(F=\{f_i\}_{i=1}^{N}\) of an image, where \(N=H\times W\) and each \(f_i\) is an individual feature at a spatial position, they are refined by the \(K\) memory items \(M=\{m_j\}_{j=1}^{K}\), where each \(m_j\) is a memory item. The read operation is based on addressing weights, which are obtained by the cosine similarities~\footnote{\textcolor{black}{The cosine similarity delivers the best performance compared to other similarity functions, such as Manhattan distance (roughly -0.8\% in terms of mIoU under the same experimental settings).}} between each individual feature \(f_i\) (for all \(i=1,...N\)) and all memory items \(\{m_j\}_{j=1}^{K}\) and a softmax function. Thus, an addressing weight \(w_{i,j}\) of an individual feature \(f_i\) to the \(j^{th}\) memory item \(m_j\) is as follows:

$$
w_{i,j} = \frac{\exp(s_{i,j})}{\sum_{k=1}^K \exp(s_{i,k})}, \eqno{(1)}
$$
where the cosine similarity \(s_{i,j}\) is computed as:

$$
s_{i,j} = \frac{f_i^T m_j}{|f_i||m_j|}. \eqno{(2)}
$$

As the proposed memory module contains a small number of memory items for compactness, each individual feature \(f_i\) addresses all items for the diverse representations instead of the one most similar item. For the feature \(f_i\), the memory module refines it through a weighted sum of all memory items with the corresponding addressing weight as:

\begin{figure}[t]
\centerline{\includegraphics[width=0.48\textwidth]{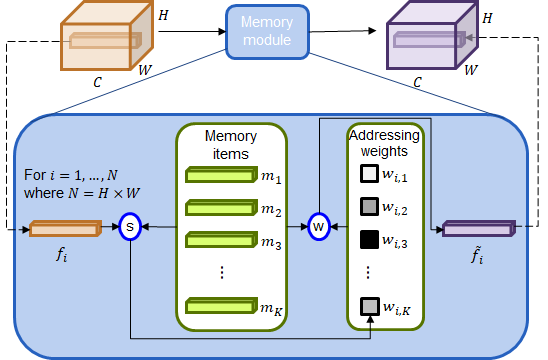}}
\caption{Illustration of the process of memory read. To read memory items, the cosine similarities of each individual feature \(f_i\) and all memory items \((m_1,...,m_K)\) are computed in (1), and then a weighted sum of the items with the corresponding addressing weight is applied as in (3) to obtain the refined feature \(\Tilde{f}_i\). \textcircled{\raisebox{-.9pt} {$\textup{s}$}} and \textcircled{\raisebox{-.9pt} {$\textup{w}$}} denote the cosine similarity operation and a weight sum.}
\label{fig:mem_read}
\end{figure}

$$
\Tilde{f}_{i} = \sum_{j=1}^K w_{i,j}m_j. \eqno{(3)}.
$$

Instead of only feeding the refined feature maps \(\Tilde{F}=\{\Tilde{f}_i\}_{i=1}^{N}\) into the decoder, they are multiplied by a scale parameter \(\gamma\) and added to the original feature maps \(F\) as \(F\) also contains the significant information of the input image. Thus, the feature maps fed into the decoder is given by:
$$
G = F+\gamma\Tilde{F}. \eqno{(4)}
$$
The parameter \(\gamma\) is a trainable scalar and initialized as 0.1.

\subsubsection{Write} The write operation is the process of updating memory items \(M=\{m_j\}_{j=1}^{K}\) using the individual features \(\{f_i\}_{i=1}^{N}\). Different from the read operation that all memory items are involved to refine each individual feature, the write operation updates each memory item using a part of individual features, each of which is most similar to the target memory item. Thus, given a target memory item \(m_j\), we first look for the features which have the highest addressing weight on \(m_j\) according to the similarities computed in (2) as:

$$
A_j=\{i\in\{1,...N\}|j=\textup{argmax}_ks_{i,k}  \textup{ 
for all } k=1,...,K\}, \eqno{(5)}
$$
where \(A_j\) contains the indexes of the features which have the highest addressing weight on \(m_j\).
Similar to (1), an update weight \(v_{j,i}\) of a memory item \(m_j\) to an individual feature \(f_i\) is computed as:

$$
v_{j,i} = \frac{\exp(s'_{j,i})}{\sum_{n=1}^N \exp(s'_{j,n})}, \eqno{(6)}
$$
where the cosine similarity \(s'_{j,i}\) is computed as:

$$
s'_{j,i} = \frac{ m_j^T f_i}{|m_j||f_i|}. \eqno{(7)}
$$
Moreover, the update weight is re-normalized by the maximum weight of the features in the set \(A_j\) as:

$$
\Tilde{v}_{j,i} = \frac{v_{j,i}}{\max_{n\in A_j}v_{j,n}}. \eqno{(8)}
$$

At last, the memory item \(m_j\) is updated by the features in the set \(A_j\) using a weighted sum with the corresponding update weight as follows:
% \(\Tilde{v}_{j,i}\)
$$
\Tilde{m}_{j} = h(m_j+\sum_{i\in A_j} \Tilde{v}_{j,i}f_i), \eqno{(9)}
$$
where \(h(\cdot)\) is L2 normalization function. If all individual features are involved in updating the memory items, update weights on similar features get diminished as small weights are assigned to the uncorrelated features. As a result, the memory items could be improperly updated. Thus, we select above updating strategy to write memory items.

\subsection{Optimization}
To train our proposed model, we exploit 2D multi-class cross entropy loss \(l_{CE}\) for semantic segmentation as:

$$
l_{CE} = \frac{1}{H_0\times W_0}\sum_{p=1}^{H_0\times W_0} -y_p\textup{log}\hat{y}_p, \eqno{(10)}
$$
where \(y_p\) and \(\hat{y}_p\) are the true category label and the segmentation predicted probability for pixel \(p\).
In addition, in order to reduce the redundancy of memory items, the triplet loss \(l_{trip}\) is used to make the items far apart as:

$$
l_{trip} = \sum_{i=1}^N \max(||f_i-m_{p_i}||_2-||f_i-m_{q_i}||_2+\alpha,0), \eqno{(11)}
$$
where \(m_{p_i}\) and \(m_{q_i}\) are the first and second most similar memory items to the feature \(f_i\) according to the addressing weights computed in (1), and \(\alpha\), set as 1.0 in our experiments, is the margin between the two items. To minimize the triplet loss, the feature \(f_i\) should be close to the \(m_{p_i}\) while far away from \(m_{q_i}\).
Thus the overall loss function \(L\) consists of the two loss functions as:

$$
L = l_{CE} + \beta l_{trip}. \eqno{(12)}
$$
\(\beta\) is set as 0.05 and the model is trained end-to-end to minimize the overall loss function.

\section{Experiments}

To evaluate the proposed method, a series of experiments are conducted. The dataset and implementation details are introduced first. Then, to look for the optimal experimental settings for achieving the best performance, some studies are delivered. At last, quantitative and qualitative results are presented.

\subsection{Dataset}
\textbf{RUGD} dataset~\cite{RUGD2019IROS} is a dataset tailored for semantic segmentation in unstructured environments. It focuses on \textit{off-road} autonomous navigation scenario. It was collected from a Clearpath Husky ground robot traversing in a variety of natural, unstructured semi-urban areas. It contains no discernible geometric edges or vanishing points, and semantic boundaries are highly irregular and convoluted. As such, off-road driving scenarios present a variety of challenges. It contains 4,759 and 1,964 images for training and testing sets, respectively. It has 24 categories including vehicle, building, sky, grass and etc. The resolution of the images is 688$\times$500. \textcolor{black}{The ablation studies are conducted on the training and testing sets.}

\textcolor{black}{\textbf{RELLIS} dataset~\cite{jiang2020rellis3d} is another dataset tailored for semantic segmentation in off-road environments. The dataset was collected on the Rellis Campus of Texas A\&M University and presents challenges to existing algorithms related to class imbalance and environmental topography. It contains 3,302 and 1,672 images for training and testing sets, respectively. It has 19 categories including fence, vehicle, rubble and etc. The resolution of the images is 1920$\times$1200. Due to the limitation of computing resources, the images were randomly cropped to 640$\times$640 during training.}

\subsection{Implementation Details}
\subsubsection{Training settings}
Following ~\cite{zhang2018context}, a poly learning rate policy is adopted. The initial learning rate is set as 0.01 and the learning rate at each iteration is the initial learning rate multiplied by \(({1-\frac{iter}{total\_iter}})^{0.9}\). The momentum and weight decay rates are set to 0.9 and 0.0001, respectively. The networks are trained with 8 mini-batch sizes per GPU using stochastic gradient descent (SGD). We set 150 epochs for training. As in existing methods, parameters in the encoder are initialized from the weights pretrained from the ImageNet~\cite{imagenet} while those in the decoder and the memory module are randomly initialized.
To avoid overfitting, data augmentation is exploited during training including horizontal flipping, scaling (from 0.5 to 2.0), and rotation (from -10$^{\circ}$ to 10$^{\circ}$).

\subsubsection{Networks}
To verify the effectiveness of our proposed memory module, it is adopted on a variety of networks, such as PSPNet~\cite{pspnet}, Deeplabv3~\cite{deeplabv3} and DANet~\cite{danet}, with various depths of encoders, such as MobileNetv2~\cite{sandler2018mobilenetv2}, ResNet18~\cite{resnet} and HRNet~\cite{hrnet}. The decoder denoted as `Upsampling' consists of a single convolutional layer producing a prediction map and a bilinear upsampling operation to resize the prediction map to the resolution of the input image for the final segmentation result. We conduct the ablation studies on Deeplabv3 with a lightweight encoder `ResNet18' with OS$=$16.

% \begin{table}[h]
% \caption{An Example of a Table}
% \label{table_example}
% \begin{center}
% \begin{tabular}{c|c}
% \hline
% \# items & mIoU\\
% \hline
% 0 & 3348\\
% 10 & 3419\\
% 20 & 3471\\
% 24 & 3507\\
% 30 & 3468\\
% 40 & 3407\\
% 60 & 3425\\
% 80 & 3403\\
% 100 & 3390\\
% \hline
% \end{tabular}
% \end{center}
% \end{table}

\begin{figure}[t]
\centerline{\includegraphics[width=0.48\textwidth]{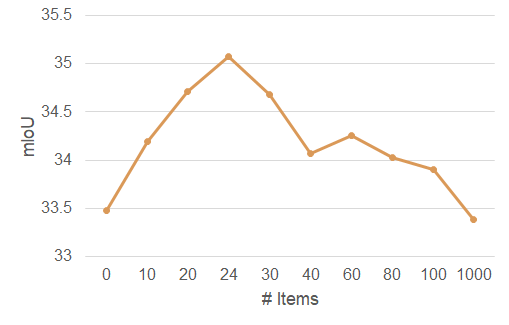}}
\caption{The effect of the number of memory items in the proposed memory module. The memory module with 24 items yields the best performance.}
\label{fig:mem_items}
\end{figure}

\subsection{Ablation Study}
\subsubsection{Effect of the Number of Memory Items}
To find the optimal number of memory items, we conduct experiments by setting the different number of memory items and the results are shown in Fig.~\ref{fig:mem_items}. We observe that the memory with 24 items, which is identical to the number of the categories, yields the best performance. It outperforms the baseline with an improvement of 1.59\% in terms of mIoU. We observe that although the memory module with less than 24 items outperforms the baseline, it cannot deliver diverse representations to cover a wide range of scenes and all categories. In the case of too many items, it is vulnerable to focus on relevant items and leads to performance degradation.

\subsubsection{Effect of triplet loss}

As the triplet loss contributes to the separateness of features, we control the separateness of the memory items to store discriminative representations by weighting the triplet loss using a constant scalar $\beta$. As shown in Table~\ref{table_example}, we vary $\beta$ from 0 to 0.2 and $\beta=0.05$ achieves the best performance.

\begin{figure}[t]
\centerline{\includegraphics[width=0.48\textwidth]{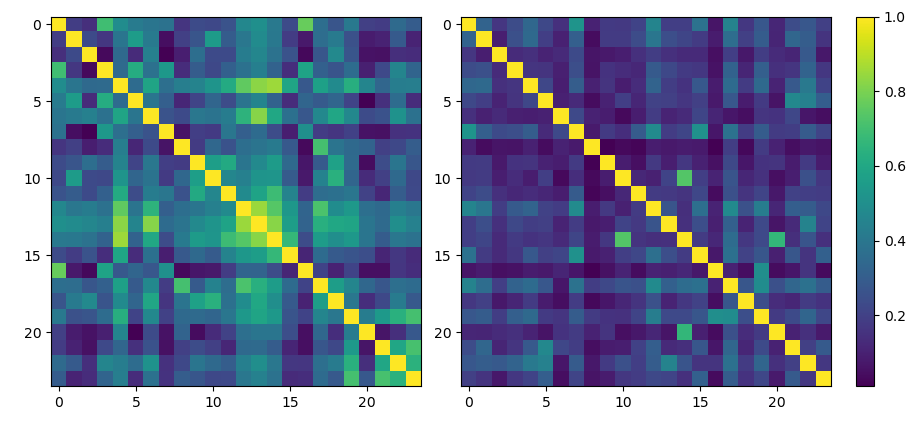}}
\caption{Visualization of cosine similarities of all pairs of memory items without (Left) and with (Right, \(\beta = 0.05\)) the triplet loss. The result demonstrates that the triplet loss makes the items highly discriminative and less redundant.}
\label{fig:mem_similarity}
\end{figure}

To analyze the effectiveness of the triple loss more clearly, we visualize the cosine similarities of all pairs of the items without/with the triple loss in Fig.~\ref{fig:mem_similarity}. We observe that the triplet loss makes the items less similar to others. The average cosine similarities of all pairs of the items without and with the triple loss are 0.50 and 0.19, respectively. These results indicate that the triplet loss allows the memory module to reduce the redundancy and store discriminative representations, which improves the segmentation accuracy.

\begin{table}[t]
\caption{Results by setting different loss weight $\beta$ on the triplet loss. Empirically, \(\beta = 0.05\) yields the best performance.}
\label{table_example}
\begin{center}
\begin{tabular}{c|ccccc}
\hline
$\beta$ & 0 & 0.01 & 0.05 & 0.1 & 0.2 \\
mIoU & 34.44 & 34.42 & 35.07 & 34.81 & 33.68\\
\hline
\end{tabular}
\end{center}
\end{table}

\subsection{Results}
To verify the effectiveness of our memory module, it is applied to diverse decoders with either compact (e.g., ResNet18 and MobileNetv2) or complex (e.g., HRNet and ResNet50) encoder. Table~\ref{tab:rugd_result} presents the segmentation performance (mIoU), the number of network parameters (\#Param) and the computational cost (GFLOPs\footnote{The GFLOPs is computed with the Pytorch code on https://github.com/sovrasov/flops-counter.pytorch.}). We can observe that our memory module can improve the performance over different networks regardless of compact or complex networks. As we propose a compact and non-parametric memory module, the baselines with our memory module keep the same number of network parameters and equivalent GFLOPs as the baseline. More importantly, ``ResNet18$\ddagger$ + Deeplabv3'' with our memory module outperforms the heavier networks ``HRNet$\dagger$ + Upsampling'' and ``ResNet50$\ddagger$ + Deeplabv3'' without the memory module. It demonstrates that our proposed memory module contributes to significant performance improvement and makes lighter networks perform as well as more complex networks.

\begin{figure}[t]
\centerline{\includegraphics[width=0.48\textwidth]{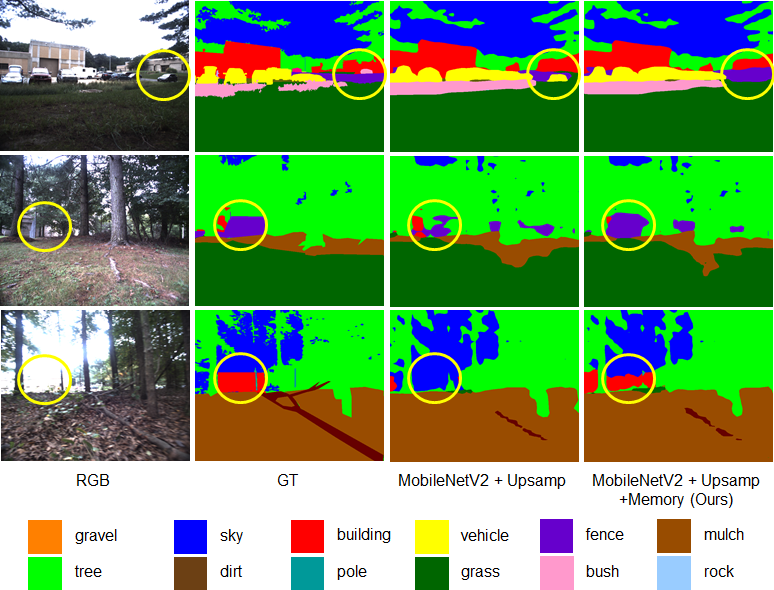}}
\caption{Visualization of semantic segmentation results from a compact network (MobileNetv2 + Upsampling). Our method is superior to capturing unclear objects, such as fences in the top two samples and a building in the bottom sample.}
\label{fig:rugd_vis_compact}
\end{figure}

\begin{figure}[t]
\centerline{\includegraphics[width=0.48\textwidth]{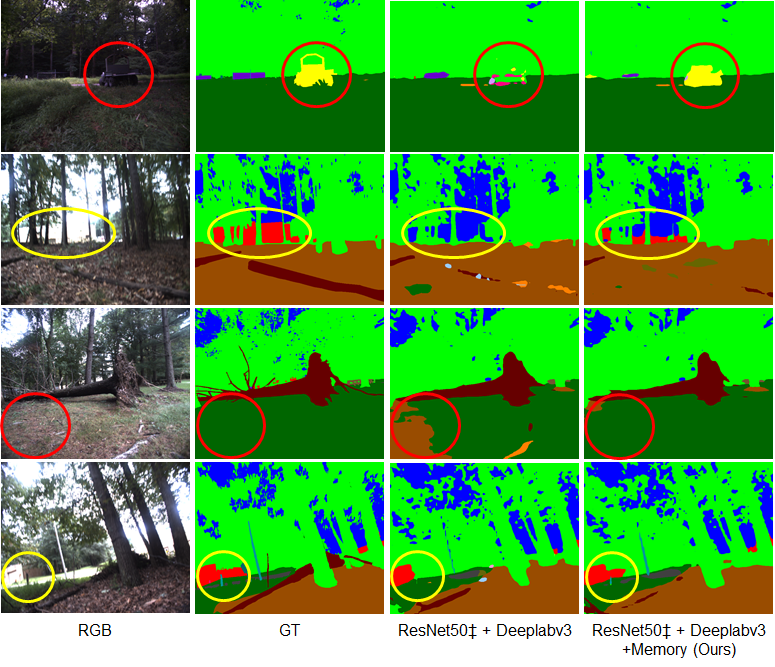}}
\caption{Visualization of semantic segmentation results from a complex network (ResNet50$\ddagger$ + Deeplabv3). Our method is superior to capturing unclear objects, such as a vehicle in the top sample, grass in the third sample and buildings in the rest of the samples.}
\label{fig:rugd_vis_complex}
\end{figure}

Fig.~\ref{fig:rugd_vis_compact} and Fig.~\ref{fig:rugd_vis_complex} give some visualization results from a compact network (MobileNetv2 + Upsampling) and a complex network (ResNet50$\ddagger$ + Deeplabv3), respectively. As shown, the results demonstrate the effectiveness of our method for capturing unclear regions and objects as highlighted regions in the images over various off-road, unstructured natural environments.

\begin{table}[t]
\caption{The segmentation results comparison with and without our memory module on the RUGD test set. `(Ours)' denotes the network with the memory module. $\dagger$ and $\ddagger$ denotes the encoder with OS$=$4 and 8 (else OS$=$16). The GFLOPs are computed on the input size 688 $\times$ 550.}
\label{tab:rugd_result}
\begin{center}
\begin{tabular}{ll|ccc}
\hline
Encoder & Decoder & mIoU & GFLOPs & \#Param \\
\hline
\hline
MobileNetv2 & Upsampling & 32.12 & 4.65 & 2.64M  \\
MobileNetv2 & Upsampling (Ours) & 32.78 & 4.66 & 2.64M \\
\hline
ResNet18 & Upsampling & 32.20 & 23.90 & 12.96M \\
ResNet18 & Upsampling (Ours) & 33.30 & 23.92 & 12.96M \\
\hline
HRNet$\dagger$ & Upsampling & 36.49 & 137.48 & 65.86M \\
HRNet$\dagger$ & Upsampling (Ours) & 37.23 & 137.74 & 65.86M \\
\hline
ResNet18 & PSPNet & 33.42  & 27.83 & 16.77M \\
ResNet18 & PSPNet (Ours) & 34.13 & 27.85 & 16.77M \\
\hline
ResNet18 & Deeplabv3 & 33.48 & 28.60 & 16.5M \\
ResNet18 & Deeplabv3 (Ours) & 35.07 & 28.62 & 16.5M \\
\hline
ResNet18 & DANet & 33.02 & 25.61 & 13.27M \\
ResNet18 & DANet (Ours) & 33.99 & 25.64 & 13.27M \\
\hline
ResNet18$\ddagger$ & Deeplabv3 & 35.98 & 93.37 & 16.5M \\
ResNet18$\ddagger$ & Deeplabv3 (Ours) & 37.04 & 93.43 & 16.5M \\
\hline
ResNet50$\ddagger$ & Deeplabv3 & 36.77 & 242.94 & 42.13M \\
ResNet50$\ddagger$ & Deeplabv3 (Ours) & 37.71 & 243.21 & 42.13M \\
\hline
\end{tabular}
\end{center}
\end{table}

\subsection{RELLIS}
\begin{figure}[t]
\centerline{\includegraphics[width=0.48\textwidth]{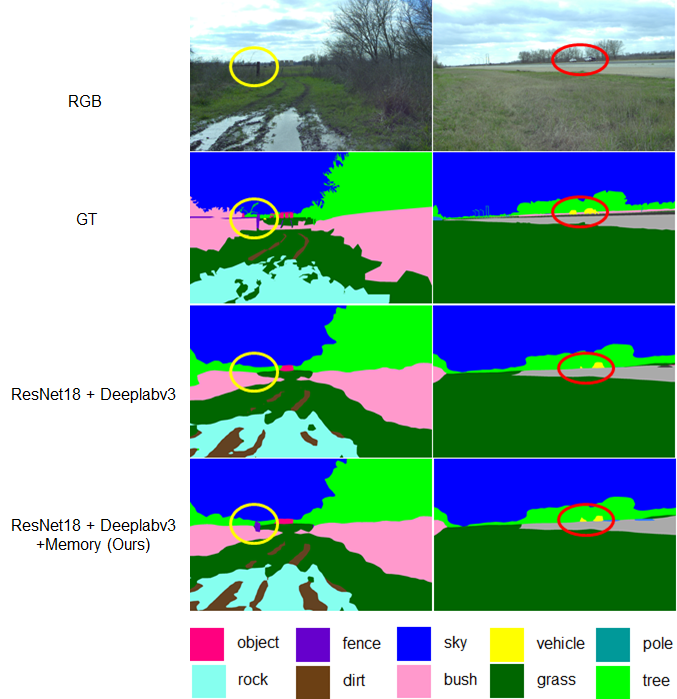}}
\caption{Visualization of semantic segmentation results from the network ``ResNet18 + Deeplabv3''. Our method is superior to capturing distant objects, such as the fence-post in the left sample and distant vehicles (especially the one on the left) in the right sample.}
\label{fig:rellis_result}
\end{figure}

\textcolor{black}{The effectiveness of our method is further validated on the RELLIS dataset. Compared to the RUGD dataset, the RELLIS dataset does not contain frame sequences with significant illumination changes.  Thus the overall quality of images captured is better than that of the RUGD dataset.  However, the RELLIS images contain scenes of wide unobstructed views, resulting in distant objects captured by a small number of pixels. As such, accurate semantic segmentation of such objects is difficult.
Table~\ref{tab:rellis_result} summarizes the test results on a variety of networks. It clearly demonstrates that our method resulted in improvement on each of the network tests. Visualization results from the network ``ResNet18 + Deeplabv3'' on RELLIS testing images are shown in Fig.~\ref{fig:rellis_result}. While the network without the memory module has difficulties accurately segmenting the fence-post and the distant vehicles (especially the one on the left), the network with our proposed memory module accurately segmented those distant objects.
}

\begin{table}[t]
\caption{The segmentation results comparison with and without our memory module on the RELLIS test set. `(Ours)' denotes the network with the memory module. $\ddagger$ denotes the encoder with OS$=$8 (else OS$=$16).}
\label{tab:rellis_result}
\begin{center}
\begin{tabular}{ll|c}
\hline
Encoder & Decoder & mIoU \\
\hline
\hline
\hline
MobileNetv2 & Upsampling & 37.26  \\
MobileNetv2 & Upsampling (Ours) & 37.89 \\
\hline
MobileNetv2 & Deeplabv3 & 38.67 \\
MobileNetv2 & Deeplabv3 (Ours) & 39.24 \\
\hline
ResNet18 & PSPNet & 38.52 \\
ResNet18 & PSPNet (Ours) & 39.97 \\
\hline
ResNet18 & DANet & 38.92 \\
ResNet18 & DANet (Ours) & 40.25 \\
\hline
ResNet18 & Deeplabv3 & 38.66 \\
ResNet18 & Deeplabv3 (Ours) & 40.10 \\
\hline
ResNet18$\ddagger$ & Deeplabv3 & 40.76 \\
ResNet18$\ddagger$ & Deeplabv3 (Ours) & 41.62 \\
\hline
ResNet50$\ddagger$ & Deeplabv3 & 43.97 \\
ResNet50$\ddagger$ & Deeplabv3 (Ours) & 45.61 \\
\hline
\end{tabular}
\end{center}
\end{table}

\section{CONCLUSIONS}

In this paper, a built-in memory module was proposed to improve the semantic segmentation performance on off-road unstructured natural environments by refining global contextual feature maps. The memory module stored the significant representations of the training images as memory items. Then, the memory items were recalled to cluster instances of the same class together within the learned embedding space even when there were significant variances in embedded features from the encoder. Thus, the memory module contributed to handling the unexpected illumination changes which made objects unclear.
Considering real-time navigation of an autonomous platform, the memory module contains a small number of memory items in order not to affect the computational cost (GFLOPs). To make the best use of the memory module, the triplet loss was employed to minimize redundancy, and the memory module stored discriminative representations. 
We demonstrated the effectiveness of the proposed memory module by applying it to several existing networks. It improved performance while rarely affecting efficiency, and the qualitative results showed that our memory module contributed to capturing unclear objects over various off-road, unstructured natural environments. As the proposed method can be integrated into compact networks, it presents a viable approach for resource-limited small autonomous platforms.

\addtolength{\textheight}{-1cm}   % This command serves to balance the column lengths
                                  % on the last page of the document manually. It shortens
                                  % the textheight of the last page by a suitable amount.
                                  % This command does not take effect until the next page
                                  % so it should come on the page before the last. Make
                                  % sure that you do not shorten the textheight too much.

%%%%%%%%%%%%%%%%%%%%%%%%%%%%%%%%%%%%%%%%%%%%%%%%%%%%%%%%%%%%%%%%%%%%%%%%%%%%%%%%

%%%%%%%%%%%%%%%%%%%%%%%%%%%%%%%%%%%%%%%%%%%%%%%%%%%%%%%%%%%%%%%%%%%%%%%%%%%%%%%%

%%%%%%%%%%%%%%%%%%%%%%%%%%%%%%%%%%%%%%%%%%%%%%%%%%%%%%%%%%%%%%%%%%%%%%%%%%%%%%%%

% \section*{ACKNOWLEDGMENT}

% This material is based upon work supported by the Air Force Office of Scientific Research under award number FA2386-19-1-4001.

%%%%%%%%%%%%%%%%%%%%%%%%%%%%%%%%%%%%%%%%%%%%%%%%%%%%%%%%%%%%%%%%%%%%%%%%%%%%%%%%

\bibliographystyle{bib_dir/IEEEtran}
\bibliography{bib_dir/IEEEexample, MemRUGD}

% \begin{thebibliography}{99}

% \bibitem{c1} G. O. Young, ÒSynthetic structure of industrial plastics (Book style with paper title and editor),Ó 	in Plastics, 2nd ed. vol. 3, J. Peters, Ed.  New York: McGraw-Hill, 1964, pp. 15Ð64.
% \bibitem{c2} W.-K. Chen, Linear Networks and Systems (Book style).	Belmont, CA: Wadsworth, 1993, pp. 123Ð135.
% \bibitem{c3} H. Poor, An Introduction to Signal Detection and Estimation.   New York: Springer-Verlag, 1985, ch. 4.
% \bibitem{c4} B. Smith, ÒAn approach to graphs of linear forms (Unpublished work style),Ó unpublished.
% \bibitem{c5} E. H. Miller, ÒA note on reflector arrays (Periodical styleÑAccepted for publication),Ó IEEE Trans. Antennas Propagat., to be publised.
% \bibitem{c6} J. Wang, ÒFundamentals of erbium-doped fiber amplifiers arrays (Periodical styleÑSubmitted for publication),Ó IEEE J. Quantum Electron., submitted for publication.
% \bibitem{c7} C. J. Kaufman, Rocky Mountain Research Lab., Boulder, CO, private communication, May 1995.
% \bibitem{c8} Y. Yorozu, M. Hirano, K. Oka, and Y. Tagawa, ÒElectron spectroscopy studies on magneto-optical media and plastic substrate interfaces(Translation Journals style),Ó IEEE Transl. J. Magn.Jpn., vol. 2, Aug. 1987, pp. 740Ð741 [Dig. 9th Annu. Conf. Magnetics Japan, 1982, p. 301].

% \end{thebibliography}

\end{document}